\documentclass[10pt,journal,compsoc]{IEEEtran}
\usepackage{graphicx}
\usepackage{epsfig}
\usepackage{amsmath}
\usepackage{amssymb}
\usepackage{subfigure}
\usepackage{chngpage}
\usepackage{tabularx}
\usepackage{array}
\usepackage{times}
\usepackage{multirow}
\usepackage[english]{babel}
\usepackage{algorithm}
\usepackage{algorithmicx}
\usepackage{mathptmx} 
\usepackage{placeins}
\usepackage{booktabs}

\ifCLASSOPTIONcompsoc
\usepackage[nocompress]{cite}
\else
\usepackage{cite}
\fi
\ifCLASSINFOpdf
\else
\fi

\begin{document}
	\title{A hybrid approach of interpolations and CNN to obtain super-resolution}

	\author{Ram~Krishna~Pandey, ~\IEEEmembership{Member,~IEEE} and~A~G~Ramakrishnan,~\IEEEmembership{Senior~Member,~IEEE}% <-this % stops a space
		\IEEEcompsocitemizethanks{\IEEEcompsocthanksitem Ram Krishna Pandey and A G Ramakrishnan are with the Indian Institute of Science, Bangalore India 560012 E-mail: (ramp@iisc.ac.in and agr@iisc.ac.in) \protect\\
		}% <-this % stops an unwanted space
	}
	
	%\markboth{IEEE TRANSACTIONS ON Pattern analysis and machine  Intelligence,~Vol.~xx, No.~x, May~2018}{}

	\IEEEtitleabstractindextext{%
		\begin{abstract}
			We propose a novel architecture that learns an end-to-end mapping function to improve the spatial resolution of the input natural images. The model is unique in forming a nonlinear combination of three traditional interpolation techniques using the convolutional neural network. Another proposed architecture uses a skip connection with nearest neighbor interpolation, achieving almost similar results. The architectures have been carefully designed to ensure that the reconstructed images lie precisely in the manifold of high-resolution images, thereby preserving the high-frequency components with fine details. We have compared with the state of the art and recent deep learning based natural image super-resolution techniques and found that our methods are able to preserve the sharp details in the image, while also obtaining comparable or better PSNR than them. Since our methods use only traditional interpolations and a shallow CNN with less number of smaller filters, the computational cost is kept low. We have reported the results of two proposed architectures on five standard datasets, for an upscale factor of 2. Our methods generalize well in most cases, which is evident from the better results obtained with increasingly complex datasets. For 4-times upscaling, we have designed similar architectures for comparing with other methods.
			
		\end{abstract}
		
		\begin{IEEEkeywords}
			Super-resolution, convolutional neural networks, SISR, residual connections, natural image, PSNR, SSIM, bicubic, bilinear, nearest neighbour, interpolation, skip connection.
	\end{IEEEkeywords}}
	
	\maketitle
	
	\IEEEdisplaynontitleabstractindextext
	
	\IEEEpeerreviewmaketitle
	\IEEEraisesectionheading{\section{Introduction}\label{sec:introduction}}
	
	\IEEEPARstart{T}{he} creation of a high resolution (HR) image from one or more low resolution (LR) images is called image super-resolution (SR). Thus, SR can be classified into two broad classes: (i) Multi-image super-resolution (MISR) (ii) Single image super-resolution (SISR). MISR requires multiple input images with sub-pixel misalignment to construct a single HR image. SISR uses a database of LR and HR patches to learn a mapping function, which can then be used to construct a HR image from any (single) LR image. This technique is called example based super-resolution.
	
	Spatial resolution refers to how finely we can see the details of an image. Most of the super-resolution techniques reported in the literature deal with natural images. The effectiveness of these methods has been generally measured in terms of peak signal to noise (PSNR) and structural similarity index (SSIM). Most of these methods have shown that the down-sampled version of an image can easily be super-resolved close to the high-resolution image manifold of the original images. An algorithm or model is robust, only if it generalizes well in all the cases. For example, when we directly pass an image rather than its down-sampled version, or if the output of the model is once again fed to the model, it should result in an image of better or the same quality. 
	%    Image super-resolution can be categorized into two distinct classes (i) Multi-image super-resolution (ii) Single image super-resolution. Multiple image super-resolution requires multiple images with sub-pixel shift to obtain high-resolution images. During testing, single image super-resolution requires a single image to construct the high-resolution image ~\cite{}. 
	
	We focus on upscaling natural images by a factor of 2. We have trained both of our models on a 91-image dataset same as that used in~\cite{srcnn14} and for testing, we have used 5 standard datasets with a total of 319 images.The results show that our models generalize well for different datasets and are comparable to the recently reported super-resolution techniques in terms of PSNR, when the complexity of the dataset increases. 
	
	\section{Survey of relevant literature}
	\label{Survey of relevant literature}
	
	Interpolations like bicubic, bilinear and nearest neighbor can upscale any image to double its size and can be further applied to get more upscaling, but each has its advantages and disadvantages~\cite{cubic,linear,survey1}. These methods give rise to artifacts like pixelization, jagged contours, and over-smoothing~\cite{atifactfreeupscaling}. Bi-cubic interpolation uses 16 neighboring pixels to find the interpolant; it cannot recover fine details and the reconstruction is smoother than bilinear. Bilinear interpolation uses a weighted combination of 4 neighboring pixels. It removes visual artifacts in the reconstruction caused by using non-integral upscaling factor~\cite{artifactfreeinterpolation}. Nearest neighbor interpolation simply repeats the pixel four times. It gives piecewise constant value and hence provides sharpness in the reconstruction. The order in which the smoothness decreases is bicubic, bilinear and nearest neighbor. However, these methods are computationally very efficient. 
	
	The task of natural image super-resolution is highly ill-posed. So, some kind of prior is required to solve this problem~\cite{sisrbenchmark}. For example, one can assume as in~\cite{neighborembeding} that the image patches in the low-resolution space (manifold) have the same local geometry as the patches in the high-resolution space. This property can be exploited to fit each patch in the LR space using weighted least square in the neighborhood and the same weights can be used to reconstruct the patches in the high-resolution space. Freeman et al. \cite{freeman} assumed these patches to be in a compact representation called the dictionary. Since we need to map from the low to the high-resolution space, we need two dictionaries to capture sufficient information about the LR and the corresponding HR spaces. This task can be formulated as an optimization problem as proposed in SISR~\cite{yang2010,yang2012}, which is NP-hard. Thus, exact solution is not possible and it needs to be approximated to solve the problem. In the above work, the authors have used an intermediate representation $\alpha$ of the patch, which is assumed to be the same for both LR and HR images in the corresponding LR and HR dictionaries. The learned representation $\alpha$ is in the concatenated feature space of LR and HR and can be assumed as a one-layer transformation to find a common representation for patches in both the spaces.
	
	Timofte et al. \cite{anr,a_plus} proposed anchored regression-based techniques, where they divided larger dictionaries into smaller ones to make the algorithms computationally faster. In~\cite{srcnn14,srcnn16}, Dong et al. use bicubic interpolation to go to the high-resolution space and then pass the output to a neural network that has multiple layer representations of the input interpolated image and reconstruct the HR image. The authors show improvement in terms of PSNR over bicubic interpolation and SISR~\cite{yang2010,yang2012}. In FSRCNN \cite{FSRCNN}, the authors propose an improvement over SRCNN by learning a mapping function without using bicubic interpolation. They use smaller size filters and more depth with transposed convolution at the last layer to upscale the images and show that the reconstructions are faster with high restoration quality. In VDSR~\cite{vdsr}, the authors show an improvement over SRCNN using smaller filters with a depth of 20 layers and learning the residual with gradient clipping and with a learning rate 104 times that of SRCNN.
	
	In \cite{DRCN}, Kim et al. propose a deeply-recursive convolutional network, recursive-supervision and skip-connection to ease the training process, while avoiding vanishing and exploding gradient problems. LapSRN~\cite{LapSRN} proposes a deep Laplacian pyramid super-resolution network and uses transposed convolution~\cite{deconv} in place of bicubic and trains the network with deep supervision using Charbonnier loss functions utilizing recursive layer to share the parameters. Other improvements in terms of computation and qualitative and quantitative results have been proposed in~\cite{perceptualloss} using perceptual and content losses together with MSE to show that the fine details are preserved a little more. A recent development in natural image SR is SRGAN~\cite{srgan}, which uses GAN~\cite{gan} with VGG nets~\cite{vggnet} and perceptual loss~\cite{perceptualloss}. The method reconstructs an image near the high-resolution, natural image manifold, which looks perceptually sharper even if it has low PSNR and SSIM.
	
	We preserve the fine details by expanding the input space using different features from the input image using different interpolation approaches. We then pass these features to a CNN (which is light weight, unlike other deep learning based techniques that use large number of parameters) to learn the mapping function from the low-resolution to the high-resolution manifold.
	
	The important point is that in all the techniques proposed in the literature, we need to reach the high-resolution image manifold preserving the structure from the low-resolution image manifold, which is the assumption in most of the cases \cite{neighborembeding}. This can be achieved through multiple ways, such as interpolation, transposed convolution \cite{deconv} or sub-pixel convolution~\cite{subpixel}. In the current work, we focus on interpolation techniques coupled with CNN and skip connection by the nearest neighbor to reconstruct the HR image. 
	
	We have proposed two main architectures. In the first architecture, we extract the features from the input image, interpolate these features by different interpolations, concatenate and pass them to the CNN as shown in Fig.~\ref{arch_2_2}. The second uses skip connection with the nearest neighbor to reconstruct the HR image as shown in Fig.~\ref{arch_2_3}. We performed independent experiments with skip connections of each of the three interpolations and found that skip connection by the nearest neighbor interpolation gives better PSNR than those achieved with bicubic or bilinear interpolation. Both of our proposed architectures for an upscaling factor of 2 are comparable to all the traditional and recent deep learning based techniques and outperform them in terms of PSNR when the complexity of the dataset increases.
	
	Most of the work in the literature take a single channel or gray scale image to obtain super-resolution~\cite{anr,neighborembeding,freeman,a_plus,srcnn14,yang2010,yang2012}. We have approached to solve this problem by taking all the color channels (RGB) together, similar to~\cite{srcnn16, dai_rgb, multichannel}.
	
	\section{Issues addressed and contributions}
	\label{Issues addressed}
	In this work, we have addressed the problem of natural image super-resolution in a different way to ensure that the proposed methods are not optimized to perform well for a particular dataset, but rather generalize well for many datasets and are also computationally more efficient. Given a low-resolution natural image, we want to obtain a high-resolution counterpart, which preserves the fine details in the reconstructed image, such that even the smaller objects in the image are distinctly seen. This can be formulated mathematically as follows. Suppose that the given low-resolution image is $I_{n}^{lr}$, the task is to learn a mapping function $\theta$ to obtain a high-resolution image $I_{n}^{hr}$ from it. $\theta(I_{n}^{lr}) \approxeq I_{n}^{hr}$ should be similar to the ground truth image (can be compared in terms of PSNR), if $I_{n}^{lr}$ was obtained by some method of downsampling. The output image, $\theta(I_{n}^{lr})$ should be better than the input image in terms of perceptual quality, if the ground truth is not available.
	
	We have exploited the traditional interpolation techniques and the convolutional neural network to obtain the high resolution image. The input images are taken to the high-resolution space by one or more interpolation methods, which can be viewed as creating diverse datasets to cover the input image space. The convolution block can be viewed as the one, which assigns weights to these interpolation techniques. In other words, the filters are learned in such a way that the interpolation that contributes better to the output pixel is given a higher weight than the others.
	Our main contributions are as follows: 
	\begin{itemize}
		\item We have designed multiple architectures for upscaling natural images by factors of 2 and 4.  
		\item  We have proposed judicious combinations of well-known interpolations, employed smaller filter sizes and shallow structures and obtained lightweight, effective architectures.
		\item Our models use less number of parameters, the details of which are mentioned in Table~\ref{parameter comparison}. The maximum number of convolution layers in any of our architectures is five.
		\item Our algorithm~\ref{CI2 algorithm} can be thought as encompassing the input image space three times compared to other techniques such as~\cite{srcnn14,srcnn16}.
		\item The simple nearest neighbor (NN) interpolation, not normally considered useful for this application, has been shown by us to play a major role in the generation of the HR image, including a local skip connection of NN.
		\item 
		We have learnt an end-to-end mapping function, where the artifacts (such as aliasing, ringing, edge halo and frequency artifact) that might have been created by the various interpolations are removed in the final output.
		
	\end{itemize}
	
	\section{Proposed SR architectures}
	\label{Proposed superresolution architecture}
	
	\subsection{Initial Experiments} 
	\label{4_1}
	We have progressively trained and obtained multiple models. As learning experiments, we initially trained models individually with only bicubic, bilinear or nearest neighbor interpolations with CNN, of which the first is similar to that proposed in~\cite{srcnn14,srcnn16}. For all these experiments, the test images from Set5 or Set14 are downsampled by two using bicubic interpolation and used as input. For all the models, all the three (RGB) channels are input directly. We found that all these techniques can deliver a reasonable quality of image reconstruction, but each method contributes more to some pixels than others in the reconstructed image. Tables~\ref{table_2_1} to ~\ref{table_2_6} list the results of the initial experiments, before arriving at the idea of combining the results obtained by all these interpolations. Table~\ref{table_2_1} shows Bicubic-CNN has better PSNR and SSIM. Table~\ref{table_2_2} shows both PSNR and SSIM is better for NN-CNN. Tables~\ref{table_2_2} to~\ref{table_2_6} show that the SSIM of bicubic-CNN is less than the Bilinear-CNN and NN-CNN. In Table~\ref{table_2_4}, SSIM of Bilinear CNN is better than NN-CNN. Overall, the results in the tables~\ref{table_2_1} to~\ref{table_2_6} show that each method contributes more to some pixel than the others in constructing  the output image and hence an optimal weighting of all these techniques or features extracted from them may give better generalized results. 
	
	\begin{table}
		\caption {Mean PSNR (in dB) and SSIM values obtained by the initial experiments on set5 images for 2X upscaling. Inputs are downsampled test images blurred by Gaussian blur of kernel size $5\times5 $.}
		
		%\makegapedcells
		%\centering 
		{\resizebox{0.48\textwidth}{0.05\textwidth}
			{ \begin{tabular}{| c| c c c |  }
					\hline
					\multirow{2}{*}{Metric} & \multicolumn{3}{c|}{Upscale by 2 of Gaussian blurred images from set5}  \\[1pt] \cline{2-4}
					&  Bicubic-CNN &  Bilinear-CNN & NN-CNN \\[3pt]
					\hline
					PSNR  & 32.19 & 31.78 & 32.06  \\[1pt]

					SSIM & 0.84 & 0.83 & 0.83  \\[1pt]	
					\hline
					
				\end{tabular}%
		}}
		\label{table_2_1}
	\end{table}
	
	\begin{table}
		\caption {Results of the initial experiments on set5 images, after 2X upscaling. Inputs are down-sampled test images, with no blur applied. The output images are smoothed using a bilateral filter.}
		
		%\makegapedcells
		%\centering 
		{\resizebox{0.48\textwidth}{0.05\textwidth}
			{ \begin{tabular}{| c|c c c   |}
					\hline
					\multirow{2}{*}{Metric} & \multicolumn{3}{c|}{Upscale by 2 of set5 images; output images smoothed.}  \\[1pt] \cline{2-4}
					& Bicubic-CNN &  Bilinear-CNN &  NN-CNN \\[1pt]
					\hline
					PSNR  & 30.96 & 31.07 & 31.52  \\[1pt]

					SSIM  & 0.76 & 0.79 & 0.82  \\[1pt]
					
					\hline
					
				\end{tabular}%
		}}
		\label{table_2_2}
	\end{table}
	
	\begin{table}
		\caption {Mean PSNR (in dB) and SSIM values obtained by the initial experiments on set5 images. Downsampled test images are directly fed as input, without any blurring.}
		
		%\makegapedcells
		%\centering 
		{\resizebox{0.48\textwidth}{0.05\textwidth}
			{ \begin{tabular}{| c | c c c  |  }
					\hline
					\multirow{2}{*}{Metric} & \multicolumn{3}{c|}{Upscale by 2 of set5 images. }  \\[1pt] \cline{2-4}
					&  Bicubic-CNN &  Bilinear-CNN &  NN-CNN \\[1pt]
					\hline
					PSNR  & 29.23 & 29.40 & 30.11  \\[1pt]

					SSIM & 0.54 & 0.59 & 0.71 \\[1pt]
					
					\hline
					
				\end{tabular}%
			}
		}
		\label{table_2_3}
	\end{table}
	\begin{table}
		\caption {Mean PSNR (in dB) and SSIM values obtained by the learning experiments on test set14, after 2X upscaling. Inputs are down-sampled test images blurred by Gaussian blur of kernel size $5\times5 $.}
		
		%\makegapedcells
		%\centering 
		{\resizebox{0.48\textwidth}{0.05\textwidth}
			{ \begin{tabular}{| c|c c c  | }
					\hline
					\multirow{2}{*}{Metric} & \multicolumn{3}{c|}{Upscale by 2 of Gaussian-blurred images from set14}  \\[1pt] \cline{2-4}
					&  Bicubic-CNN &  Bilinear-CNN &  NN-CNN  \\[1pt]
					\hline
					PSNR & 31.32 & 31.07 & 31.28  \\[1pt]

					SSIM & 0.76 & 0.76 & 0.74 \\[1pt]		
					\hline
					
				\end{tabular}%
		}}
		\label{table_2_4}	
	\end{table}
	
	\begin{table}
		\caption {Mean PSNR (in dB) and SSIM values obtained by the initial experiments on set14 test images after 2X upscaling. The reconstructed output image is smoothed with a bilateral filter.}
		
		%\makegapedcells
		%\centering
		{\resizebox{0.48\textwidth}{0.05\textwidth} 
			{ \begin{tabular}{| c | c c c  | }
					\hline
					\multirow{2}{*}{Metric} & \multicolumn{3}{c|}{Set14 upscaled by 2; Output images bilateral smoothed}  \\[1pt] \cline{2-4}
					&  Bicubic-CNN &  Bilinear-CNN &  NN-CNN  \\[1pt]
					\hline
					PSNR & 30.34 & 30.45 & 30.9  \\[1pt]

					SSIM & 0.73 & 0.73 & 0.76  \\[1pt]	
					\hline
					
				\end{tabular}%
		}}
		\label{table_2_5}	
	\end{table}
	
	\begin{table}
		\caption {Mean PSNR and SSIM values obtained by the initial experiments on set14 test images after 2X upscaling.  Test images are directly input to the model, without any blurring. }
		
		%\makegapedcells
		\centering 
		{\resizebox{0.48\textwidth}{0.05\textwidth}
			{ \begin{tabular}{| c| c  c  c  |  }
					
					\hline
					\multirow{2}{*}{Metric} & \multicolumn{3}{c|}{Upscale by 2 of set14 test images with no blur.}  \\[1pt] \cline{2-4}
					&  Bicubic-CNN & Bilinear-CNN &  NN-CNN   \\[1pt]
					\hline
					PSNR  & 28.90 & 29.05 & 29.65  \\[1pt]

					SSIM & 0.48 & 0.53 & 0.65  \\[1pt]
					
					\hline
					
				\end{tabular}%
		}}
		\label{table_2_6}
	\end{table}
	\subsection{Our proposed algorithms/architectures}
\begin{figure*}
		\centering
		\includegraphics[width=0.95\textwidth,height=0.27\textheight]{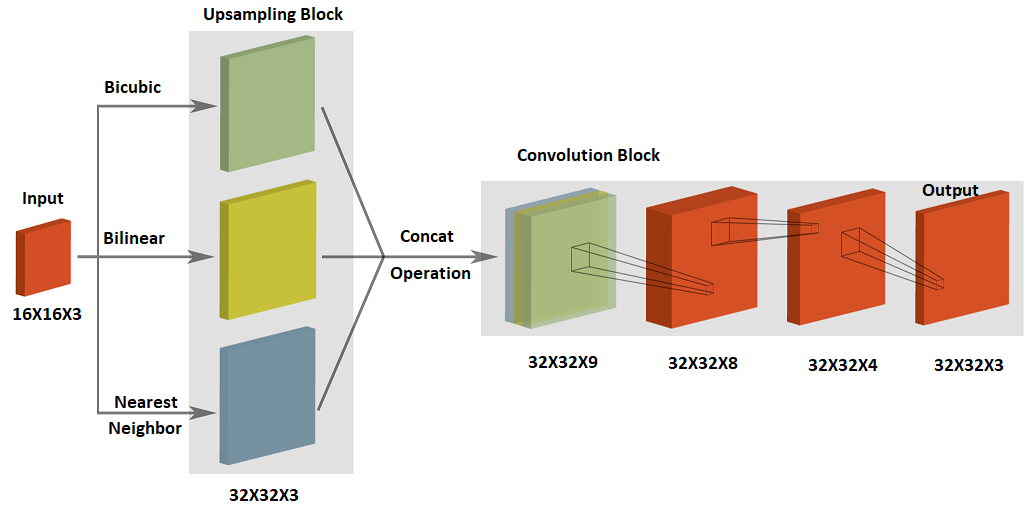}
		\caption{Interpolation by two and convolution (I2C) architecture used for upscaling by a factor of 2. Three different interpolations, each by a factor of 2, are combined by a convolutional neural network to obtain superresolution.}
		\label{arch_2_1}
	\end{figure*}

	Algorithm~\ref{I2C algorithm} (I2C), shown as an architecture diagram in Fig.~\ref{arch_2_1}, gives an overview of our approach to combine multiple interpolations in a deep architecture. Here we directly pass the interpolated output after concatenation to the CNN.  The idea is that, whichever method contributes better in reconstructing the output pixel, should be given more weightage. Also, using the three interpolations helps us to expand the input image space 3 times the case, when a single interpolation method is used.
	
\begin{figure*}
		\centering
		\includegraphics[width=0.95\textwidth,height=0.27\textheight]{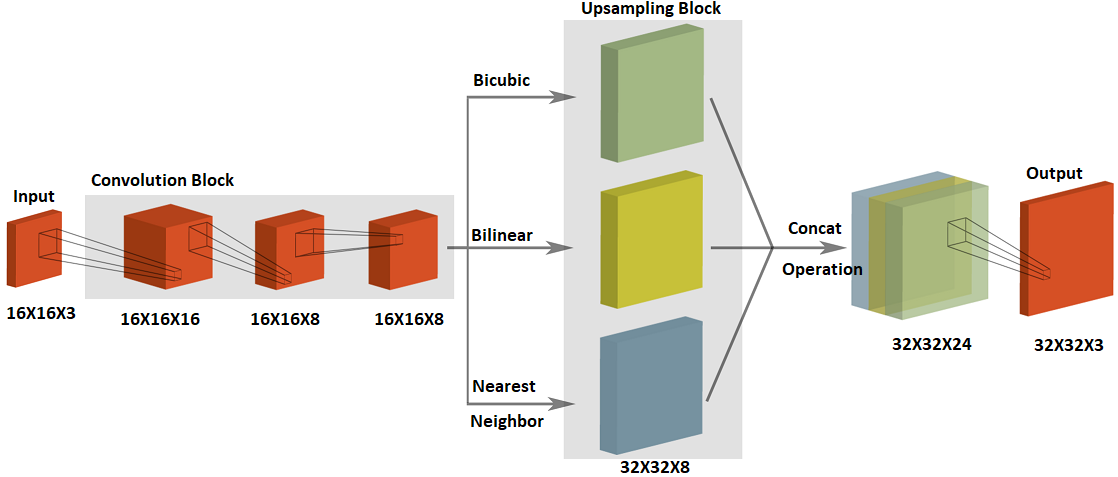}
		\caption{Convolution- interpolation by two (CI2) architecture used for upscaling by a factor of 2. This architecture is an improvement over our I2C architecture. Extraction of features before interpolation helps improve the quality of the final generated image.}
		\label{arch_2_2}
	\end{figure*}

	Algorithm~\ref{CI2 algorithm} (CI2), shown as an architecture diagram in Fig.~\ref{arch_2_2}, is an improvement over algorithm \ref{I2C algorithm}. Extracting the features before passing to the various interpolations results in an output image better than the case, when we first perform the interpolations and pass all the outputs to the convolution block. Algorithms~\ref{I2C algorithm} and~\ref{CI2 algorithm} show that extracting CNN features before performing interpolations is better than passing the concatenated, interpolated outputs to the CNN.
	
	\begin{algorithm}
		
		1. $ \bold {Input:}$ \hspace{0.5cm} $ I_{k}^{lr} $
		
		2. $ \bold {Initialize:}$ \hspace{0.05cm} $ Y_{0} = I_{k}^{lr} $
		
		3. \hspace{1cm}$ I_{bic} = bicubic(Y_{0})$
		
		4. \hspace{1cm}$ I_{bil} = bilinear(Y_{0})$
		
		5. \hspace{1cm}$ I_{nn} = nearest neighbor(Y_{0})$
		
		6. \hspace{1cm}$Z_{0}$ = $ [ I_{bic}, I_{bil} , I_{nn} ]$;
		
		7. \hspace{1cm}where, [] represents the concatenation operation (CO) 
		
		8. \hspace{.2cm}$\bold{for} $ $ \hspace{0.2cm} i =1:3 $ $\bold{do}$	
		
		9. $\hspace{1cm} \zeta^{j}_{i} =W^{j}_{i}*Z_{i-1} $ ;
		where $*$ denotes the convolution.
		
		10. $\hspace{0.85cm}Z_{i}= \max(\zeta_{i}^{j} , 0)\hspace{.5cm}  $ 
		
		11. $\bold{end} $
		
		12. $\bold{Output:}$$\theta_{\lambda}(I_{k}^{lr}) = Z_{3} $ 
		
		\caption{:I2C algorithm (architecture is shown in Fig.~\ref{arch_2_1}).}
		\label{I2C algorithm}
		
	\end{algorithm}
	
	\begin{algorithm}
		1. $  \bold {Input:}$ \hspace{0.5cm} $ I_{k}^{lr} $
		
		2. $  \bold {Initialize:}$ \hspace{0.5cm} $Z_{0}$= $I_{k}^{lr} $
		
		3.  $ \bold{ for} \hspace{0.2cm} i =1:4 $ $\bold{do}$
		
		4. $ \hspace{2cm} \zeta^{j}_{i} =W^{j}_{i}*Z_{i-1} $ ;
		
		5. \hspace{1cm}$ \hspace{1cm}Z_{i}= \max(\zeta_{i}^{j} , 0)\hspace{.5cm}  $ 
		
		6. \hspace{1cm}$\bold{if} \hspace{0.2cm} i = = 3\hspace{0.2cm} \bold{do}$
		
		7. \hspace{2cm}  $ I_{bic} = bicubic(Z_{3})$
		
		8. \hspace{2cm} $ I_{bil} = bilinear(Z_{3})$
		
		9. \hspace{2cm} $ I_{nn} = nearest neighbor(Z_{3})$
		
		10. \hspace{1.9cm}$Z_3$ = $ [ I_{bic}, I_{bil} , I_{nn} ]$; where [] is CO 
		
		11. \hspace{0.7cm} $\bold{end}$  
		
		12. $\bold{end} $
		
		13 $\bold{Output:}$ $\theta_{\lambda}(I_{k}^{lr}) = Z_{4} $ 
		
		\caption{:CI2 algorithm (architecture is shown in Fig.~\ref{arch_2_2}).}
		\label{CI2 algorithm}
	\end{algorithm}
	%\vspace{-0.1cm}
	
\begin{figure*}
		\centering
		\includegraphics[width=0.95\textwidth,height=0.27\textheight]{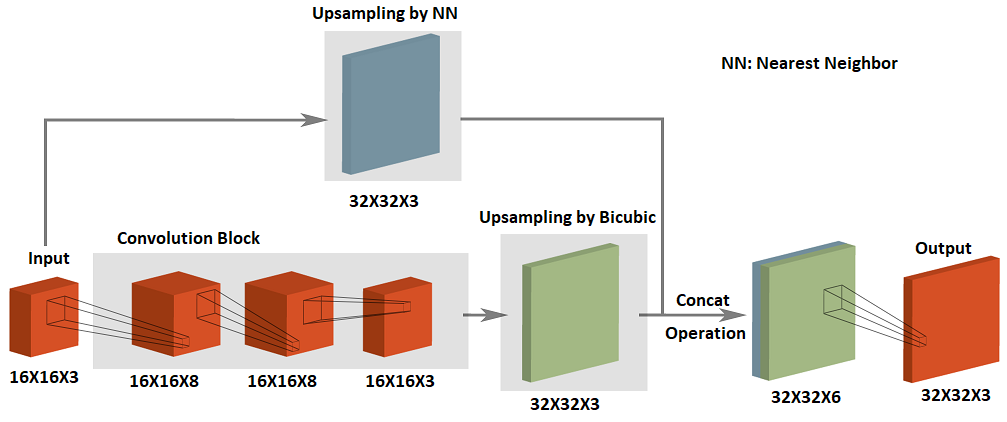}
		\caption{Convolution-bicubic interpolation by two- skip NN (CB2SNN) architecture used for upscaling by 2. This architecture uses skip connection by nearest neighbor together with bicubic upsampling to achieve an output image  comparable that given by the CI2 architecture.}
		\label{arch_2_3}
	\end{figure*}

	Algorithm~\ref{CB2SNN algorithm} (CB2SNN), shown as an architecture diagram in Fig.~\ref{arch_2_3}, uses a local skip connection by nearest neighbor. This is motivated by the initial experiments in Sec.~\ref{4_1}, which clearly show that NN-CNN has an edge over the other two interpolations followed by CNN. This CB2SNN architecture obtains PSNR comparable to that of our CI2 algorithm (Algorithm 2). 
	
	\begin{algorithm}
		1. $\bold{Input:}$ $ I_{k}^{lr} $ \\
		2. $\bold{Initialize:}$ $ Z_{0} = I_{k}^{lr} $\;
		
		3. \hspace{1cm} $ I_{nn} = nearest neighbor(Z_{0})$
		
		4. $\bold{ for} \hspace{0.2cm} i =1:4 \hspace{.2cm} \bold {do}$	
		
		5. $ \hspace{1cm} \zeta^{j}_{i} =W^{j}_{i}*Z_{i-1} $ ;
		where $*$ denotes the convolution.
		
		6. $ \hspace{1cm}Z_{i}= \max(\zeta_{i}^{j} , 0)\hspace{.5cm}  $ 
		
		7.  \hspace{1cm}$\bold {if} \hspace{0.2cm} i = = 3  \hspace{0.2cm} \bold {do}$
		
		8.  \hspace{2cm} $ I_{bic} = bicubic(Z_{3})$
		
		9. \hspace{2.1cm}$Z_{3}$ = $ [ I_{bic}, I_{nn} ]$; where [] is CO 
		
		10.  \hspace{0.7cm} $\bold {end}  $	
		
		11. $\bold{end} $
		
		12. $\bold{Output:}$ $\theta_{\lambda}(I_{k}^{lr}) = Z_{4} $		
		\caption{:CB2SNN algorithm (Fig.~\ref{arch_2_3} shows the architecture).}
		\label{CB2SNN algorithm}
	\end{algorithm}
	%\vspace{-0.1cm}
	
\begin{figure*}
		\centering
		\includegraphics[width=0.95\textwidth,height=0.22\textheight]{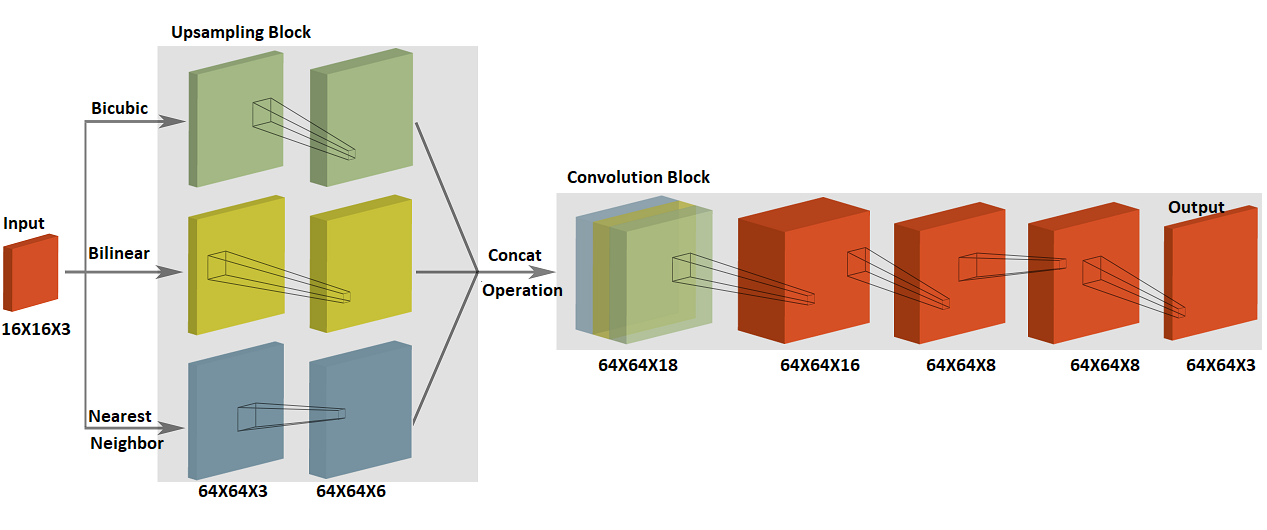}
		\caption{Interpolation-by-four - convolution (I4C) architecture proposed for superresolution by a factor of 4. This I4C architecture is somewhat similar to the I2C architecture in Fig. 1, but has an independent convolution stage for each of the interpolation outputs before concatenating them.}
		\label{arch_4_1}
	\end{figure*}

	Algorithm~\ref{I4C algorithm}, shown in the architecture form in Fig.~\ref{arch_4_1}, is used to upscale any input natural image by a scale factor of 4, similar to that used for an upscale factor of 2 given in Fig.~\ref{arch_2_2}. We directly interpolate the input image 4 times by the three interpolations and then pass the concatenated tensor to the convolutional block.
	\begin{algorithm}
		1. $\bold{Input:}$ $ I_{k}^{lr} $
		
		2. 	$\bold{Initialize:}$ $ Y_{0} = I_{k}^{lr} $
		
		3. 	\hspace{2cm}$ I_{bic4} = bicubic4(Y_{0})$
		
		4. 	\hspace{2cm}$ I_{bil4} = bilinear4(Y_{0})$
		
		5. 	\hspace{2cm}$ I_{nn4} = nearest neighbor4(Y_{0})$
		
		6. 	$ \hspace{2cm} conv_{bic} =W^{j}_{1}*I_{bic4} $ 
		
		7.	$ \hspace{2cm}Z_{bic}= \max(conv_{bic} , 0)\hspace{.5cm}  $ 
		
		8.	$ \hspace{2cm} conv_{bil} =W^{j}_{1}*I_{bil4} $ 
		
		9.	$ \hspace{2cm}Z_{bil}= \max(conv_{bil} , 0)\hspace{.5cm}  $ 
		
		10.	$ \hspace{1.9cm} conv_{nn} =W^{j}_{1}*I_{nn4} $ 
		
		11.	$ \hspace{1.9cm}Z_{nn}= \max(conv_{nn} , 0)\hspace{.5cm}  $
		
		12.	\hspace{1.9cm}$Z_{1} = [Z_{bic},Z_{bil},Z_{nn}]$; where [] is CO;
		
		13. 	$ \bold{ for} \hspace{0.2cm} i =2:5 \hspace{0.2cm} \bold {do} $	
		
		14.	$ \hspace{1.9cm} \zeta^{j}_{i} =W^{j}_{i}*Z_{i-1} $ 
		
		15.	$ \hspace{1.9cm} Z_{i}= \max(\zeta_{i}^{j} , 0)\hspace{.5cm}  $ 
		
		16.	$\bold{end} $
		
		17.	$\bold{Output:}$ $\theta_{\lambda}(I_{k}^{lr}) = Z_{5} $ 
		
		%		\vspace{.05cm}
		\caption{:I4C algorithm (architecture is shown in Fig.~\ref{arch_4_1}).}
		\label{I4C algorithm}
	\end{algorithm}
	
	\begin{figure*}
		\centering
		\includegraphics[width=0.95\textwidth,height=0.22\textheight]{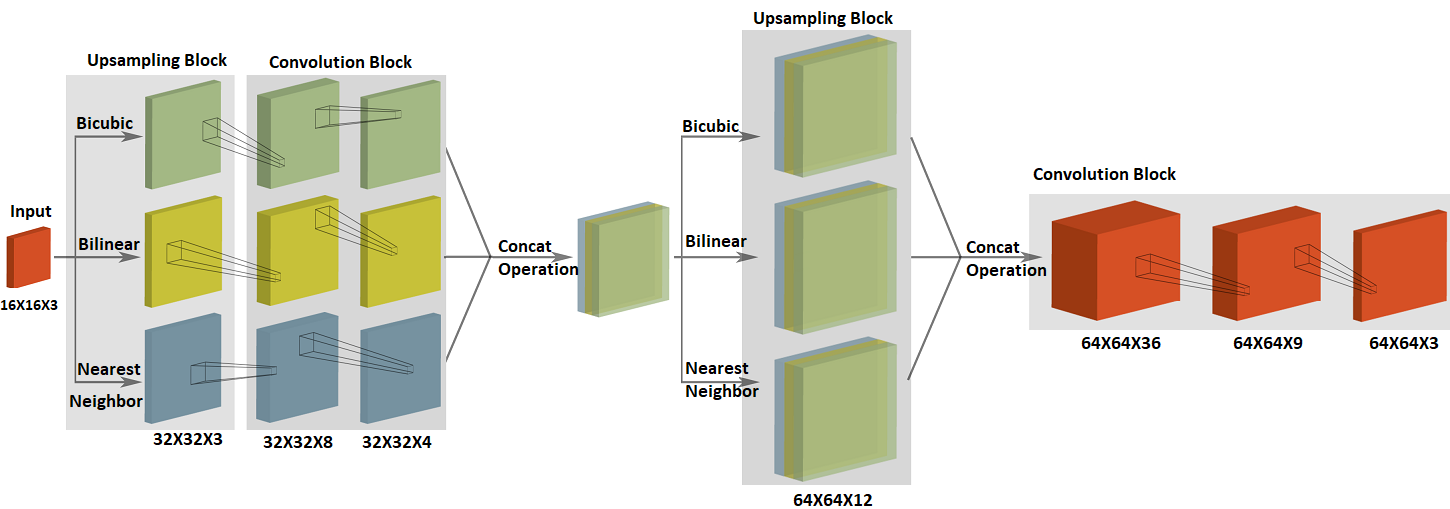}
		\caption{Interpolation by two-convolution-interpolation by two-convolution (I2CI2C) architecture for superresolving by a factor of 4. This architecture combines the ideas of I2C and CI2 architectures proposed for upscaling by a factor of 2.}
		\label{arch_4_2}
	\end{figure*}
	
	Algorithm~\ref{I2CI2C algorithm} (I2CI2C), shown as an architecture diagram in Fig.~\ref{arch_4_2}, is similar to the algorithm ~\ref{CI2 algorithm} used for an upscale factor of 2. Here, we first interpolate by a factor of 2 and pass the outputs parallely to CNNs. The outputs of the CNNs are concatenated and again interpolated by a factor of 2 by all the three interpolation schemes. These interpolated outputs are again concatenated and then passed to a two-layer CNN to get the final HR output.
	
	\begin{algorithm}
		1. $\bold{Input:}$ $I_{k}^{lr} $
		
		2. $\bold{Initialize:}$ $ Y_{0} = I_{k}^{lr} $
		
		3. \hspace{2cm}$ Zbc_{0} = bicubic(Y_{0})$
		
		4. \hspace{2cm}$ Zbl_{0} = bilinear(Y_{0})$
		
		5. \hspace{2cm}$ Znn_{0} = nearest neighbor(Y_{0})$
		
		6. $ \bold{ for} \hspace{0.3cm} i =1:2\hspace{0.2cm} \bold {do} $	
		
		7. $ \hspace{2cm} conv_{bic} =W^{j}_{i}*Zbc_{i-1} $ 
		
		8. $ \hspace{2cm}Zbc_{i}= \max(conv_{bic} , 0)\hspace{.5cm}  $ 
		
		9. $ \hspace{2cm} conv_{bil} =W^{j}_{i}*Zbl_{i-1} $ 
		
		10. $ \hspace{1.9cm}Zbl_{i}= \max(conv_{bil} , 0)\hspace{.5cm}  $ 
		
		11. $ \hspace{1.9cm} conv_{nn} =W^{j}_{i}*Znn_{i-1} $ 
		
		12. $ \hspace{1.9cm}Znn_{i}= \max(conv_{nn} , 0)\hspace{.5cm}  $ 
		
		13. $\bold{end}$
		
		14. $Z_{2} = [Zbc_{2},Zbl_{2},Znn_{2}]$; where [] is CO;
		
		15. $ \bold{ for} \hspace{0.2cm} i  =3:5 \hspace{0.2cm} \bold {do} $	
		
		16. $ \hspace{2cm} \zeta^{j}_{i} =W^{j}_{i}*Z_{i-1} $ 
		where $*$ denotes the convolution.
		
		17. $ \hspace{2cm} Z_{i}= \max(\zeta_{i}^{j} , 0)\hspace{.5cm}  $ 
		
		18. $\bold{end}  $
		
		19. $\bold{Output:}$	$\theta_{\lambda}(I_{k}^{lr}) = Z_{5} $ 
		\vspace{.1cm}
		\caption{:I2CI2C algorithm (architecture is shown in Fig.~\ref{arch_4_2}).}
		\label{I2CI2C algorithm}
		\vspace{-0.1cm}
	\end{algorithm}

	The training set is defined as $I = \{I_{k}^{lr},I_{k}^{hr}) : 1 \leq k \leq N\}$, where $I_{k}^{lr}$ are the LR patches fed to the model to reconstruct an output similar to the HR ground truth patches. The model is trained to minimize the mean square error (MSE) between the output and the corresponding ground truth images. $I_{k}^{hr} \approxeq \theta_{\lambda}(I_{k}^{lr}) $ is the corresponding HR image patch from the training set. This is a supervised learning algorithm and once the model converges, the weights are fixed and the model is saved. Then, any test set can be fed as input to construct the corresponding HR image. The model is represented as,
	$\lambda = \{ W_{i}, b_{i} \}$ where
	$ W_{i} $ = $ \big\{ W_{i}^{j} : \hspace{.1cm} 1\leq j \leq n_{i}\big\} $, and $W_{i}^{j}$ is the $j^{th}$ filter at the $i^{th}$ layer. The loss function is given by,
	\[L(\theta_{\lambda}(I^{lr}),I^{hr})=\frac{1}{N}\sum\limits_{k=1}^{N}\parallel \theta_{\lambda}(I_{k}^{lr})-I_{k}^{hr}\parallel_{F}\]
	
	To minimize this loss function, Adam optimizer is used. The choice of the parameters used for Adam optimizer are: learning rate (lr)= 1e-4, $\beta_{1}=0.9, \beta_{2}=0.999, \epsilon = 1e-8$. The details of the algorithm and the analysis of its convergence can be found in~\cite{adam}. Further refinement in the convergence analysis can be found in~\cite{onconvergenceadam}, along with the failure cases and their solution.
	
	Deep learning models can easily be represented in terms of the architectural details. Figure~\ref{arch_2_1} shows the interpolation-by-two-convolution (I2C) architecture proposed for training our model. we first interpolate the image by bicubic, bilinear and nearest neighbor methods and pass their concatenation on to a 3-layer-CNN. The CNN uses 8 filters of size $5 \times 5$ in the first layer, 4 filters of size $3\times3$ in the second layer and three filters of size $ 3 \times 3 $ in the last layer. This shows how multiple interpolation outputs can be combined to construct the HR image by incorporating multiple variations (indirectly) in the input image itself. The idea here is to give weights to each of these methods (interpolations) in such a way that the method contributing the best in constructing the output pixel is given more weight by the convolution block.
	
	Motivated by the performance of the I2C architecture (algorithm~\ref{I2C algorithm}), we have developed a new architecture (algorithm~\ref{CI2 algorithm}) shown in Fig.~\ref{arch_2_2}. In this convolution-interpolation by two (CI2) algorithm, we first pass the input image to a 3-layer convolution block to extract the best possible features. The convolution block uses 16 filters of size $ 5 \times 5 $ in the first layer, and 8 filters each of size $ 3\times 3$ in both the second and third layers. We then feed the extracted features to the upsampling block, which interpolates them using the three interpolation techniques. Their concatenated outputs are then passed on to a one-layer convolution block with 3 filters of size $ 3 \times 3$ to form the HR image. We have observed that adding more layers after the upsampling block does not further improve the reconstruction. So, we have removed them and use only one layer to achieve a low computational complexity model.
	
	Algorithm~\ref{CB2SNN algorithm} (architecture shown in Fig.~\ref{arch_2_3}) has a slightly different architecture and shows that the nearest neighbor makes a good contribution to our design. We have performed extensive experiments and found that a skip connection with the nearest neighbor gives a result better than that of a skip connection with bilinear or bicubic interpolation. This convolution-bicubic interpolation by two- skip NN (CB2SNN) architecture delivers results comparable to that of the CI2 architecture. However, it uses 3 times less number of parameters as given in Table~\ref{parameter comparison}, and is computationally more efficient than the CI2 architecture shown in Fig.~\ref{arch_2_2}. The input image is fed to the first layer of CNN with 8 filters of size $5 \times 5$, then to the second layer with 8 filters of size $3 \times 3$ and the third layer of 3 filters of size $ 1 \times 1$. The output is bicubic interpolated and  concatenated with the input interpolated by nearest neighbor and fed to the last layer of CNN with 3 filters of size $3 \times 3$.  
	
	Algorithm~\ref{I4C algorithm} (architecture shown in Fig.~\ref{arch_4_1}) gives the steps of the Interpolation-by-four-convolution (I4C) architecture, designed for upscaling the natural images by a factor of 4. We first upscale the input patch by 4 times by the three interpolation techniques, namely, nearest neighbor, bicubic and bilinear to the size $64\times64\times3$. The three outputs are fed parallely to distinct convolutional blocks, each having 6 filters of size $5\times5$, to construct the HR image patches. The features obtained after the convolutions are concatenated and are passed on to the second layer of CNN having 16 filters of size $5\times5$. The third layer has 8 filters of size $3\times3$; the fourth layer, 8 filters of size $1\times1$; and the final layer, 3 filters of size $3\times3$. This architecture gives a good result, but a better result is obtained when we follow an architecture similar to the CI2 shown in Fig.~\ref{arch_2_2} (proposed for upscaling by a factor of 2).
	
	The details of the Interpolation by two - convolution - interpolation by two - convolution (I2CI2C) architecture are shown in Fig.~\ref{arch_4_2}. The input image is first interpolated by the three interpolations to the size $32\times32\times3$. The outputs are passed parallely to the first layer of CNN containing 8 filters, each of size $3 \times 3$. The second layer uses 4 filters each of size $ 1\times1 $. We concatenate these interpolated feature maps, before upsampling again by a factor of two, parallely by all the three interpolation schemes. We again concatenate these outputs and pass them on to another CNN with 9 filters of size $3 \times 3$ and the last layer with 3 filters of size $3 \times 3$.
	
	The 4 times upscaling architectures are developed to directly upscale the input image (and not the downsampled one) by 4 times. Thus, there is no ground truth image, and it is difficult to compare the results with those of other methods in terms of PSNR. However, we have qualitatively compared with one of the recently developed deep learning based model, namely SRGAN~\cite{srgan}.
	%where the authors have used the concept of generative adversarial network~\cite{gan} together with vgg net~\cite{vggnet} model and claimed that the reconstructed image lies precisely in  the manifold of HR image. 
	We observe that the perceptual quality of our output images is comparable to that obtained by SRGAN. Our method is simple with less computational complexity than SRGAN that has VGG net, which uses millions of parameters.
	
	We arrived at our best architectures after performing extensive experiments. Figures~\ref{results_2_fig_2} and~\ref{results_2_fig_1} compare the performance of our architectures (in terms of PSNR and SSIM) with that of bicubic interpolation (Bicubic) as the baseline. We also show the results of feeding the outputs of each of these interpolations to a CNN. Very good results obtained by CI2 and CB2SNN architectures are shown in Fig.~\ref{results_2_fig_2} for a sample test image (`woman') taken from Set14 \cite{set14}. Bicubic-CNN can be considered similar to~\cite{srcnn14}. Bilinear-CNN and NN-CNN are the results obtained by the CNN operating on the bilinear and nearest neighbor interpolations of the input image, respectively. The results shown in Fig.~\ref{results_2_fig_1} are the poorest of our results among all the images in the datasets tested.
	
	\section{Training and test datasets used}
	\label{Dataset}
	While creating the training and test datasets, we have focused on upscaling the input images by a factor of 2. For training, we created the LR patches in various ways: downsampling by bicubic, bilinear, nearest and pyramid. The sizes of the LR and the corresponding HR patches are $ 16 \times 16 \times 3$ and $32 \times 32 \times 3$, for upscaling by a factor of 2. These patch pairs are sampled from 91 training images. The data is created keeping the correspondence of the patch pairs intact. Based on our multiple experiments, we have found that the inclusion of pyramid downsampled LR patches gives rise to slightly better results and generalization. The total number of patch pairs created using bicubic downsampling is 1035019 for 2X upsampling. Similarly, for upscaling by a factor of 4, the patch pairs are of size $16\times16 \times 3$ and $64\times64\times3$, sampled from the same 91 images. For testing, the five standard data sets, namely, set5 \cite{set5}, set14, BSD100 \cite{bsd100}, Urban100 \cite{urban100} and DIV2K \cite{div2k} are used, which have 5, 14, 100, 100 and 100 test images, respectively, making up a total of 319 images.
	
	\section{Experiments, Results and Discussion}
	\label{Experiments and Discussion}
	%\begin{figure*}[!ht]
	%	\centering
	%	\includegraphics[width=0.99\textwidth,height=0.30\textheight]{cvpr_lr_14_nbb_baboon.png}
	%	\caption{Results on one of the test image from the dataset set14}
	%	\label{fig: 2}
	%\end{figure*}
	
\begin{figure*}
		\centering
		\includegraphics[width=0.98\textwidth,height=0.25\textheight]{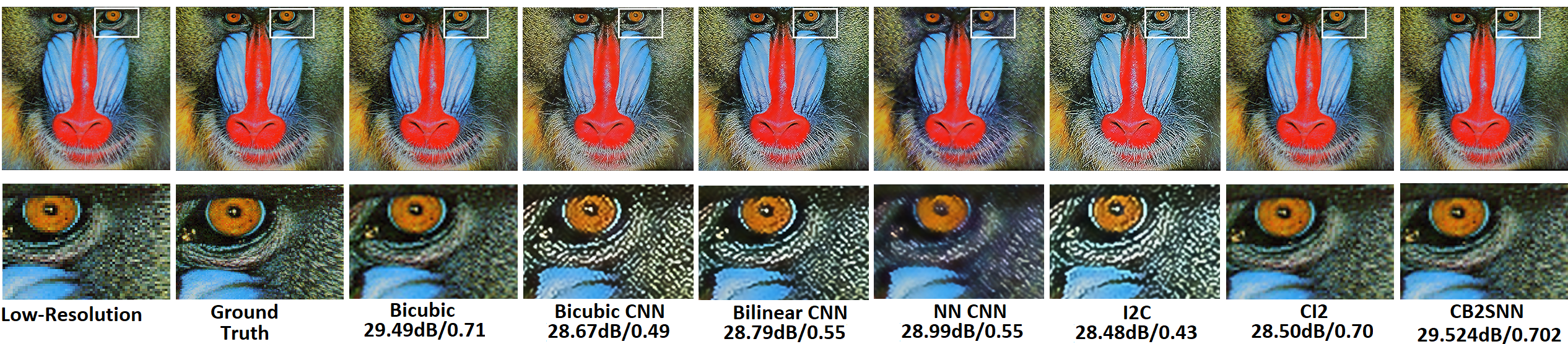}
		\caption{Comparison of the results of many experiments performed by us in doubling the resolution of 'baboon' (a test image from the dataset set14). These are the poorest results we have obtained with our methods for any test image. CB2SNN gives the best PSNR of 29.524 dB, while the simple bicubic interpolation delivers an image with comparable PSNR (29.49 dB) and SSIM values.}
		\label{results_2_fig_1}
	\end{figure*}

	Our complete results on all the five datasets are given in the supplementary material. Figures~\ref{results_2_fig_2} and \ref{results_2_fig_1} show the results for the images 'woman' from set5 and 'baboon' from set 14. Baseline results shown are reconstructions by normal bicubic interpolation and outputs of CNN models trained using bicubic (Bicubic CNN), bilinear (Bilinear CNN) and nearest neighbor (NN CNN) interpolations. The idea of combining bicubic with CNN is similar to that reported by Dong et al. ~\cite{srcnn14}, but with less number of lower sized filters. The main results reported are those obtained by our I2C, CI2 and CB2SNN algorithms, out of which the latter two achieve the best reconstructions. Thus, hereafter, we focus mainly on the results of our two best algorithms, namely CI2 and CB2SNN.
	
	\begin{figure*}
		\includegraphics[width=0.98\textwidth,height=0.23\textheight]{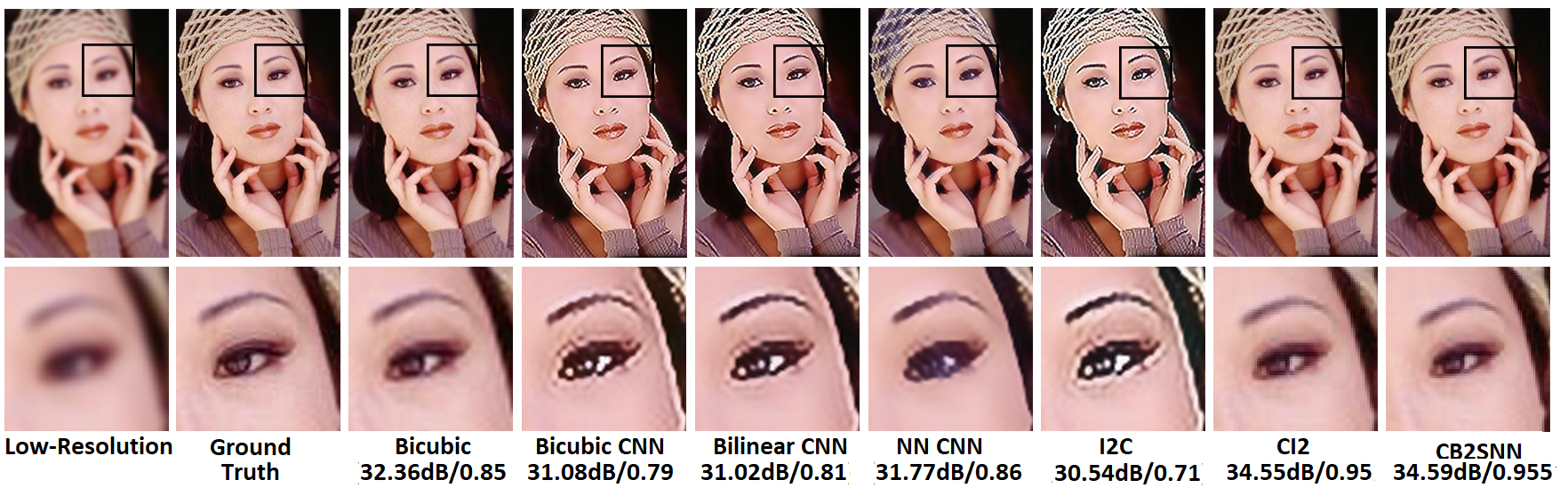}
		\caption{Comparison of the results of many experiments performed by us to obtain the best results in doubling the resolution of 'woman' (a test image from the dataset set5). The output of CB2SNN achieves the best PSNR of 34.59 dB and SSIM of 0.955. The results of CI2 are close, with a PSNR of 34.55 dB and a SSIM of 0.95, compared to a PSNR of 32.36 dB and SSIM of 0.85 by the baseline Bicubic.}
		\label{results_2_fig_2}
	\end{figure*}
	
	Tables~\ref{table_2_7} and~\ref{table_2_8} list the PSNR and SSIM of all the images and their mean, and also on separate channels of Y, Cb and Cr on the datasets Set5 and Set14 by our CI2 architecture proposed for an upscaling factor of 2. Table~\ref{table_2_9} lists the average PSNR and SSIM on the datasets BSD100, Urban100 and DIV2K. Since these datasets contain 100 images each, it is difficult to list the individual PSNR and SSIM. However, we have listed individual PSNR and SSIM for the datasets Set5 and Set14.
	
	\begin{table}
		\caption {P(PSNR in dB) and S (SSIM) values obtained by our CI2 architecture on Set5 test images down-sampled by 2 and fed as input. The subscript gives the channel for which the performance is given.}
		
		%\makegapedcells
		\centering 
		\resizebox{0.48\textwidth}{0.08\textwidth}
		{ \begin{tabular}{|| c||c c c c c c c c|| }
				\hline
				Image & $P_{RGB}$ & $S_{RGB}$ &  $P_{Y}$ &  $S_{Y}$ & $ P_{Cb}$ & $ S_{Cb} $& $P_{Cr}$ & $S_{Cr}$ \\[1pt]
				\hline
				\hline
				woman & 34.55 & 0.95 & 34.55 & 0.93 & 46.80 & 0.99 & 51.2 & 0.99 \\[1pt] 
				%\hline
				baby & 36.80 & 0.96& 36.92&0.95 & 36.92& 0.99 & 50.30 &0.99\\[1pt]
				%\hline 
				butterfly & 32.21 & 0.92& 32.21& 0.88& 43.52& 0.98& 44.61& 0.99\\[1pt]
				%\hline
				head & 32.73 & 0.83& 33.87&0.83& 37.95&0.88& 39.55&0.91\\[1pt]
				%\hline
				bird & 36.89 & 0.97& 37.10& 0.97& 45.57 & 0.99& 42.81& 0.99\\[1pt]
				\hline 
				
				\bf Mean & \bf 34.63 & \bf 0.93 & \bf 34.93& \bf 0.91& \bf 44.83& \bf 0.97& \bf 45.7 & \bf 0.97 \\[1pt]
				
				\hline			
			\end{tabular}%
		}
		\label{table_2_7}
		\vspace{-0.1in}
	\end{table}
	
	\begin{table}
		\caption {PSNR (in dB) and SSIM values obtained by our CI2 architecture on the individual color components of each of the Set14 images. Test images are down-sampled by 2, before being fed as input.} 
		%\makegapedcells
		\centering 
		\resizebox{0.48\textwidth}{0.15\textwidth}
		{ \begin{tabular}{|| c ||c c c c c c c c|| }
				\hline
				Image & $ P_{RGB} $ &  $S_{RGB}$ & $P_{Y} $ & $S_{Y}$ & $P_{Cb}$ & $S_{Cb}$ & $P_{Cr}$ & $S_{Cr}$ \\[1pt]
				\hline
				\hline
				lenna & 34.04 & 0.87 & 35.13 & 0.88 & 38.85 &  0.91  & 40.17 & 0.92 \\[1pt] 
				%\hline
				ppt3    & 35.28 & 0.94 & 35.43 &0.93  & 42.10 &  0.98  & 41.36  & 0.98\\[1pt]
				%\hline 
				man     & 32.50  & 0.85 & 32.58 & 0.81 & 52.88 &  1.00   & 53.42 & 1.00\\[1pt]
				%\hline
				barbara & 32.50 & 0.85 & 32.75 &0.84  & 43.04 &  0.97  & 42.78 & 0.97\\[1pt]
				%\hline
				comic   & 30.89 & 0.89 & 30.93 & 0.83 & 38.98 &  0.97 & 41.95 & 0.97\\[1pt]
				%\hline 
				flower  & 33.21 & 0.90 & 33.53 & 0.88 & 39.02 &  0.95 & 38.83  & 0.94 \\[1pt]
				%\hline
				zebra   & 32.96 & 0.93 & 33.00  & 0.90 & 49.71 &  0.99 & 48.98  & 0.99 \\[1pt] 
				%\hline
				foreman & 35.61 & 0.95 & 35.71 &0.94  & 45.53 &  0.99 & 46.77  & 0.99\\[1pt]
				%\hline 
				pepper  & 33.66 & 0.85 & 35.10  & 0.87 & 37.24 &  0.90 & 37.80 & 0.90\\[1pt]
				%\hline
				monarch & 36.85 & 0.96 & 37.00  &0.95  & 48.22 &  0.99 & 49.10 & 0.99\\[1pt]
				%\hline
				baboon  & 29.50 & 0.70 & 29.85 & 0.67 & 32.26 & 0.70 & 31.97 & 0.69\\[1pt]
				%\hline 
				coastguard & 31.20 & 0.83 & 31.21 & 0.81 & 52.84 & 1.00 & 51.58  & 0.99 \\[1pt]
				%\hline 
				bridge  & 31.03 & 0.84 & 31.03 & 0.79 & 52.72 & 0.99 & 53.33  & 1.00 \\[1pt]
				%\hline 
				face    & 32.68 & 0.82 & 33.83 & 0.83 & 37.91 & 0.88 & 39.48  & 0.91 \\[1pt]
				\hline 
				
				\bf Mean & \bf 33.03 & \bf 0.87 & \bf 33.36 & \bf 0.85 & \bf 43.67 & \bf 0.94 & \bf 44.10  & \bf 0.95 \\[1pt]						
				\hline			
			\end{tabular}%
		}
		\label{table_2_8}
		\vspace{-0.1in}
	\end{table}
	
	\begin{table}
		\caption {Mean PSNR (dB) and SSIM values obtained by CI2 architecture on test dataset BSD100, URBAN100 and DIV2K.}
		\centering 
		\resizebox{0.48\textwidth}{0.050\textwidth}
		{ \begin{tabular}{|| c||c c c c c c c c|| }
				\hline
				Test set & $ P_{RGB} $& $ S_{RGB} $& $ P_{Y} $ & $ S_{Y}$ & $P_{Cb}$ &$ S_{Cb}$ & $P_{Cr}$ & $S_{Cr} $\\[1pt]
				\hline
				\hline
				BSD100& 33.03 & 0.87 & 33.04 & 0.84 & 51.39 &  0.99  & 50.82 & 0.99 \\[1pt] 
				\hline
				URBAN100 & 31.76 & 0.87 & 31.79 & 0.83   & 43.01 & 0.96  & 42.70 & 0.96 \\[1pt] 
				\hline	
				DIV2K & 35.26 & 0.91 & 35.33 & 0.90 & 48.19 & 0.98  & 47.59 & 0.98 \\[1pt] 
				\hline			
			\end{tabular}	
		}
		\label{table_2_9}	
		\vspace{-0.1in}
	\end{table}
	
	Tables~\ref{table_2_10} and~\ref{table_2_11} list the PSNR and SSIM of all the images and their mean and also on separate channels Y, Cb and Cr on the datasets Set5 and Set14 obtained by our CB2SNN algorithm. Table~\ref{table_2_12} lists the average PSNR and SSIM values on the datasets BSD100, Urban100 and DIV2K.
	
	\begin{table}
		\caption {PSNR (in dB) and SSIM values of the individual color components of each test image of Set5 obtained by CB2SNN architecture.}
		
		%\makegapedcells
		\centering 
		\resizebox{0.49\textwidth}{0.07\textwidth}
		{ \begin{tabular}{|| c|| c c c c c c c c|| }
				\hline
				Image & $P_{RGB}$ & $ S_{RGB}$ & $ P_{Y}$ & $S_{Y} $&  $P_{Cb} $ & $ S_{Cb}$ & $ P_{Cr} $ & $ S_{Cr} $ \\[1pt]
				\hline
				\hline
				woman     & 34.58   & 0.96 & 34.61  & 0.93  & 46.45       & 0.99    & 50.53  & 0.99 \\[1pt] 
				%\hline
				baby      & 36.79   & 0.96 & 36.95   & 0.95   & 49.67 & 0.99   & 49.34 & 0.99\\[1pt]
				%\hline 
				butterfly & 32.28   & 0.92  & 32.33   & 0.88   & 42.92       & 0.98  & 44.04  & 0.99 \\[1pt]
				%\hline
				head      & 32.74   & 0.83 & 33.92  & 0.83  & 37.98       & 0.88    & 39.57  & 0.91\\[1pt]
				%\hline
				bird      & 36.75  & 0.97  & 37.25   & 0.97   & 43.82       & 0.98    & 42.03  & 0.99\\[1pt]
				\hline 
				
				\bf Mean & \bf 34.63 & \bf 0.93 & \bf 35.01 & \bf 0.91 & \bf 44.17& \bf 0.97& \bf 45.60 & \bf 0.97 \\[1pt]
				
				\hline			
			\end{tabular}%
		}
		\label{table_2_10}
		\vspace{-0.1in}
	\end{table}
	
	\begin{table}
		\caption {P(PSNR in dB) and S (SSIM) values of the individual color components of each test image of Set14 obtained by CB2SNN architecture.}
		
		%\makegapedcells
		\centering 
		\resizebox{0.48\textwidth}{0.15\textwidth}
		{ \begin{tabular}{|| c|| c c c c c c c c|| }
				\hline
				Image & \bf $P_{RGB}$ & \bf $S_{RGB}$ & \bf $P_Y$ & \bf $S_Y$ & \bf $P_{Cb}$ & \bf $S_{Cb}$ & \bf $P_{Cr}$ & \bf $S_{Cr}$ \\[1pt]
				\hline
				\hline
				LENNA      & 34.05     & 0.87    & 35.13   & 0.88   & 38.82   &  0.91  & 40.15   & 0.93   \\ [1pt] 
				%\hline
				PPT3 & 35.40     & 0.94    & 35.62   & 0.93    & 42.08   &  0.98  & 41.46   & 0.98  \\ [1pt]
				%\hline 
				MAN        & 32.61     & 0.87    & 32.61   & 0.82   & 53.48  &  0.99   & 53.24   & .995    \\ [1pt]
				%\hline
				BARBARA    & 32.54     & 0.86    & 32.73   & 0.84    & 43.67   &  0.97  & 43.01   & 0.97  \\ [1pt]
				%\hline
				COMIC & 30.97    & 0.89     & 31.03   & 0.84   & 38.52   &  0.96  & 41.45   & 0.97   \\ [1pt]
				%\hline 
				FLOWER & 33.26    & 0.90     & 33.61   & 0.89   & 38.61   &  0.95  & 38.76   & 0.94   \\ [1pt]
				%\hline
				ZEBRA & 33.03    & 0.93     & 33.09   & 0.90   & 48.64   &  0.99  & 48.48   & 0.99   \\ [1pt] 
				%\hline 
				FOREMAN & 35.60    & 0.95     & 35.62   &0.94    & 46.21   &  0.99  & 47.43   & 0.99   \\ [1pt]
				%\hline  
				PEPPER & 33.67    & 0.85     & 35.08   & 0.87   & 37.07   &  0.89  & 37.76   & 0.90   \\ [1pt]
				%\hline
				MONARCH & 36.85    & 0.96     & 37.09   &0.95    & 46.60   &  0.99  & 48.10   & 0.99   \\ [1pt]
				%\hline 
				BABOON & 29.52    & 0.70     & 29.86   & 0.67   & 32.31   & 0.71   & 32.03   & 0.70   \\ [1pt]
				%\hline 
				COASTGUARD & 31.23    & 0.83     & 31.22   & 0.80   & 53.61   & 0.99  & 51.67   & 0.99   \\ [1pt]
				%\hline 
				BRIDGE & 31.03    & 0.84     & 31.03   & 0.79   & 52.83   & 0.99   &  52.80  & 0.99   \\ [1pt]
				%\hline 
				FACE & 32.72    & 0.83     & 33.85   & 0.83   & 37.94   & 0.88   & 39.50  & 0.91   \\ [1pt]
				\hline 
				\hline
				\bf MEAN & \bf 33.03 & \bf 0.87 & \bf 33.40 & \bf 0.85 & \bf 43.60 & \bf 0.94 & \bf 43.99  & \bf 0.95 \\[1pt]						
				\hline			
			\end{tabular}%
		}
		\label{table_2_11}	
		\vspace{-0.1in}
	\end{table}
	
	\begin{table}
		\caption {P (Mean PSNR in dB) and S (SSIM) values obtained by CB2SNN architecture on BSD100,URBAN100 and DIV2K test dataset.}
		
		%\makegapedcells
		\centering 
		\resizebox{0.48\textwidth}{0.050\textwidth}
		{ \begin{tabular}{|| c|| cc c c c c c c|| }
				\hline
				Test Set & \bf $P_{RGB}$ & \bf $S_{RGB}$ & \bf $P_Y$ & \bf $S_Y$ & \bf $P_{Cb}$ & \bf $S_{Cb}$ & \bf $P_{Cr}$ & \bf $S_{Cr}$ \\[1pt]
				\hline
				\hline
				BSD100 & 33.06 & 0.87 & 33.08 & 0.84   & 50.62 &  0.99  & 50.39 & 0.99 \\[1pt] 
				\hline	
				URBAN100 & 31.82 & 0.87 & 31.85 & 0.83   & 42.98 & 0.96  & 42.69 & 0.96 \\[1pt] 
				\hline	
			
				DIV2K & 35.26 & 0.92 & 35.34 & 0.90   & 47.44 & 0.98  & 47.19 & 0.98 \\[1pt] 
				\hline		
			\end{tabular}%
		}
		\label{table_2_12}
		\vspace{-0.1in}
	\end{table}
	
\begin{figure*}
		%	\centering
		\includegraphics[width=0.99\textwidth,height=0.25\textheight]{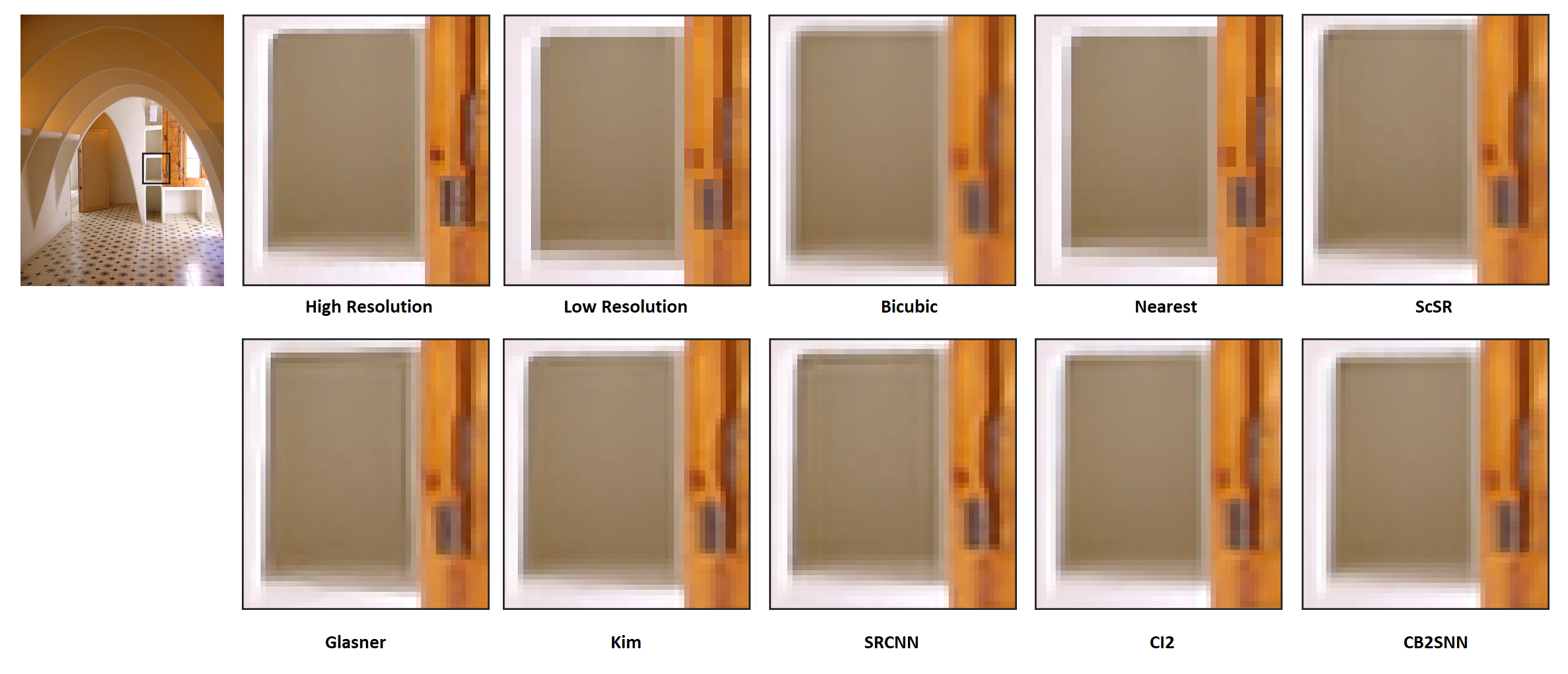}
		\caption{Comparison of the results of our 2X upscaling methods (CI2 and CB2SNN) with those of some of the techniques in the literature~\cite{yang2010,srcnn14,glasner,vdsr} on an image from the test dataset BSD100.}
		\label{results_2_fig_3}
	\end{figure*}

\begin{figure*}
		%	\centering
		\includegraphics[width=0.98\textwidth,height=0.25\textheight]{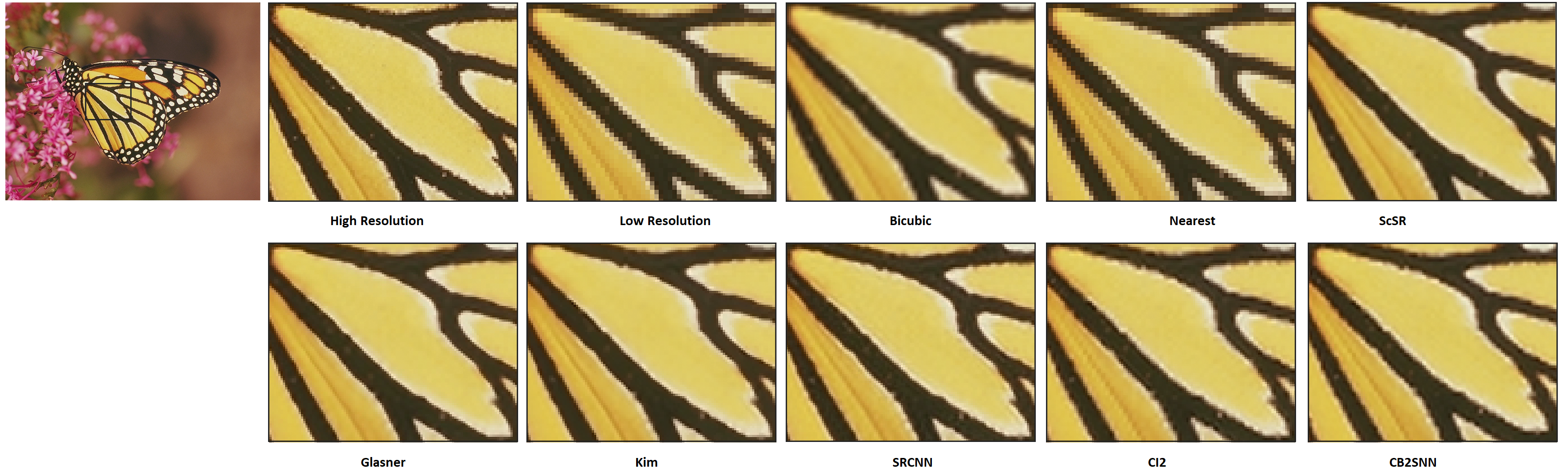}
		\caption{Comparison of the results of our 2X upscaling methods (CI2 and CB2SNN) with those of some of the techniques in the literature~\cite{yang2010,srcnn14,glasner,vdsr} on an image from the test dataset Set14.}
		\label{results_2_fig_4}
	\end{figure*}

\begin{figure*}
		%	\centering
		\includegraphics[width=0.99\textwidth,height=0.25\textheight]{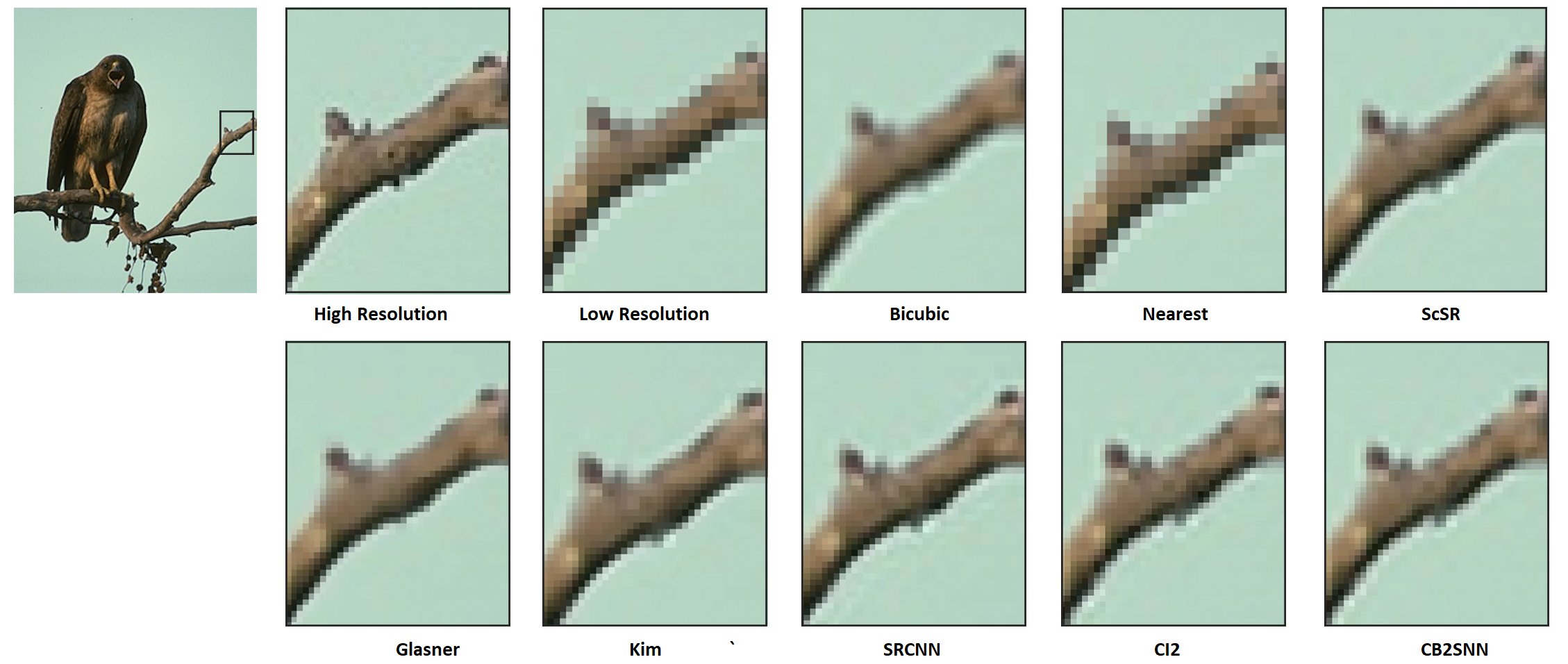}
		\caption{Comparison of the results of our 2X upscaling methods (CI2 and CB2SNN) with those of some of the techniques in the literature~\cite{yang2010,srcnn14,glasner,vdsr} on an image from the test dataset urban100. }
		\label{results_2_fig_5}
	\end{figure*}

	Figures~\ref{results_2_fig_3} to~\ref{results_2_fig_5} qualitatively compare the results of CI2 and CB2SNN architectures with those obtained by various traditional techniques together with some of the recently proposed, state of the art deep learning based techniques on one image each from the datasets BSD100, Set14, and Urban100, respectively. Table~\ref{table_2_13} compares our methods with various traditional and recently proposed deep learning based models in terms of the mean PSNR and SSIM values for each of the 5 datasets. Both CI2 and CB2SNN methods are comparable in terms of PSNR and outperform the other techniques, when the complexity of the dataset increases.
	
	\begin{table*}
		\caption {Comparison of traditional and deep learning based SR methods with the proposed CI2 and CB2SNN models in terms of mean PSNR (in dB) and SSIM on the datasets: Set5, Set14,BSD\_100,Urban\_100 and DIV2K. The test images are down-sampled by 2 to be used as input to the models.}
		
		%\makegapedcells
		\centering 
		\resizebox{0.95\textwidth}{0.12\textwidth}
		{ \begin{tabular}{|| c || c c c c c c c c c c c|| }
				\hline
				\bf Dataset& \bf Scale & \bf Metric & \bf Bicubic & \bf Lanczos3 & \bf SRCNN~\cite{srcnn14} & \bf FSRCNN~\cite{FSRCNN} & \bf VDSR~\cite{vdsr} & \bf DRCN~\cite{DRCN} & \bf LapSRN~\cite{LapSRN} & \bf CI2 & \bf CB2SNN\\[1pt]
				
				\hline
				\hline
				\multirow{2}{*}{\bf Set5} & \multirow{2}{*}{2}& PSNR& 33.66 & 34.32 & 36.66 & 37.05   & 37.53 & 37.63  & 37.52 & \bf 34.63 & \bf 34.63 \\[1pt] 
				
				& & SSIM & 0.93 & 0.94 & 0.95 & 0.96 & 0.96 & 0.96  & 0.96 & 0.93 & 0.93 \\[1pt] 
				\hline
				
				\multirow{2}{*}{\bf Set14} & \multirow{2}{*}{2}   &   PSNR & 30.24 & 30.69 & 32.45 & 32.66   & 33.03 & 33.04  & 33.08 & \bf 33.03 & \bf 33.03 \\[1pt] 
				
				& & SSIM & 0.87 & 0.88 & 0.91 & 0.91        &  0.91 & 0.91 & 0.91 & 0.87 & 0.87\\[1pt] 
				\hline	
				
				\multirow{2}{*}{\bf BSD100} & \multirow{2}{*}{2}   &  PSNR   & 29.56 & 29.92 & 31.36 & 31.53   & 31.90 & 31.85  & 31.80 & \bf 33.03 & \bf 33.06\\[1pt] 
				
				& & SSIM & 0.84 & 0.85 & 0.89 & 0.89          &  0.90 & 0.89  & 0.89 & 0.87  & 0.87 \\[1pt] 
				\hline
				
				\multirow{2}{*}{\bf Urban100} & \multirow{2}{*}{2} & PSNR   & 26.88 & 27.25 & 29.50 & 29.88   & 30.76 & 30.75  & 30.41 & \bf 31.76 & \bf 31.82 \\[1pt] 
				
				& & SSIM & 0.84 & 0.85 & 0.89 & 0.90 & 0.91 & 0.91 & 0.91 & 0.86 &0.92 \\[1pt] 
				\hline	
				
				\multirow{2}{*}{\bf DIV2K} & \multirow{2}{*}{2}    & PSNR    & 31.01 & -- & 33.05 & --   & 33.66 & --  & -- & \bf 35.26 & \bf 35.25\\[1pt] 
				
				& & SSIM & 0.94 & -- & 0.96 & --   & 0.96 & --  & -- & 0.91 & 0.92 \\[1pt] 
				\hline	
			\end{tabular}%
			
		}
		\label{table_2_13}	
	\end{table*}	
	
	\subsection{Computational complexity of the new architectures}
	Unlike other deep learning based techniques, we have tried to use less number of parameters to make our model less complex and hence faster. The computational complexity of the CNN model is discussed in ~\cite{cnncost}. The time complexity of a CNN can be written as $\mathcal{O}(\sum_{l=1}^{d}n_{l-1}.s_{l}^2.n_{l}.m_{l}^2)$, where $l$ is the index of the convolutional layer, d is the depth of convolution, $ n_{l}$ is the number of filters in the $l^{th}$ layer, $s_{l}$ is the spatial size of the filter, and  $m_{l}$ is the spatial size of the output feature maps. Table~\ref{parameter comparison} lists the number of parameters, filters and depth of our architectures and compares them with those of some deep learning based techniques.
	
	\begin{table}[!htbp]
		
		\caption {Comparison of various SR techniques for (4x)~\cite{fasr} with our techniques (2x and 4x) in terms of No. of filters (NF), network depth (ND) and number of parameters (NP). LR: low resolution; bic: bicubic; bil: bilinear; nn: nearest neighbor interpolations.}
		\centering
		\resizebox{0.48\textwidth}{0.16\textwidth}{
			\begin{tabular}{||c ||c| c | c |  c || }
				
				\hline
				\centering
				\bf Method & \bf Input & \bf NF & \bf ND & \bf  NP \\[1pt]
				\hline
				\hline
				SRCNN & LR+bic	 		& 64 & 3  & 57k  \\[1pt]
				
				FSRCNN & LR			 	& 56 & 8  & 12k \\[1pt]
				
				ESPCN & LR 			    & 64 & 3  & 20k \\[1pt]
				
				SCN & LR+bic 			& 128 & 10  & 42k \\[1pt]
				
				VDSR & LR+bic 			& 64 & 20  & 665k \\[1pt]
				
				DRCN & LR+bic			& 256 & 20  & 1775k  \\[1pt]
				
				DRRN & LR+bic			& 128 & 52  & 297k  \\[1pt]
				
				MDSR & LR 					& 64	&  162 & 8000k \\[1pt]
				
				LapSRN & LR                 & 64 & 24  & 812k \\[1pt]
				
				\bf I2C & LR+bic+bil+nn 		& 15 & 3  & 2k \\[1pt]
				
				\bf CI2 & LR+bic+bil+nn		& 35 & 4  & 3k \\[1pt]
				
				\bf CB2SNN & LR+bic+nn		& 22 & 4  & 1k  \\[1pt]
				
				\bf I4C & LR+bic+bil+nn 		& 53 & 5  & 9.7k \\[1pt]
				
				\bf I2CI2C & LR+bic+bil+nn 		& 52 & 4  & 3.8k  \\[1pt]
				\hline
				
			\end{tabular}
			\label{parameter comparison}
		}
		
	\end{table}
	
	Keeping the above mentioned time complexity analysis into consideration, we have designed the architectures such that our model complexity is as low as possible. We have not used more than 5 convolution layers in any of our architectures. The maximum filter size is $5\times 5$, and the maximum size of the output feature maps are $32\times32$ and $64\times64 $ in the architectures used for upscaling by a factor of 2 and 4, respectively.
	
	\subsection{Discussion}
	In algorithm~\ref{I2C algorithm}, we attempted a weighted combination of the three interpolation methods. However, the obtained results indicated that we could obtain better reconstruction when we combine these techniques in the best possible way, which led to developing the algorithm~\ref{CI2 algorithm}. Here, we have used more layers to extract features and then passed them on to the CNN and found that more details can be preserved. We performed detailed experiments to assign weights to each of these interpolations before feeding them to the CNN. We have found that when we give equal weights to all the interpolations, the results are better in terms of PSNR and SSIM. 
	
	After confirming that the use of multiple interpolations with a CNN can lead to better results, we shifted our focus to get better-generalized results for different inputs. As can be seen in Fig.~\ref{arch_2_2}, instead of passing the input image directly, we extracted more complex features (using multiple layers) from the input by the use of a CNN and then we have used the upsampling block to upsample the features and concatenate the interpolated features and pass it further to a one-layer CNN to reconstruct the output. Our aim is better generalization with reduced computational cost while preserving the quality. Multiple layers after the concatenated features in Figure~\ref{arch_2_2} can give slightly better results in terms of PSNR but we have used only one layer to reduce the computation. The robustness of our methods can be seen in the results. Our methods perform well in terms of PSNR for all the datasets used, except set5. The results are comparable to almost all the recently developed super-resolution techniques for natural images.
	
	\subsection{Results of super-resolution by a factor of 4}
	\label{Results of X4}
	
\begin{figure*}
		%\centering
		\includegraphics[width=0.99\textwidth,height=0.20\textheight]{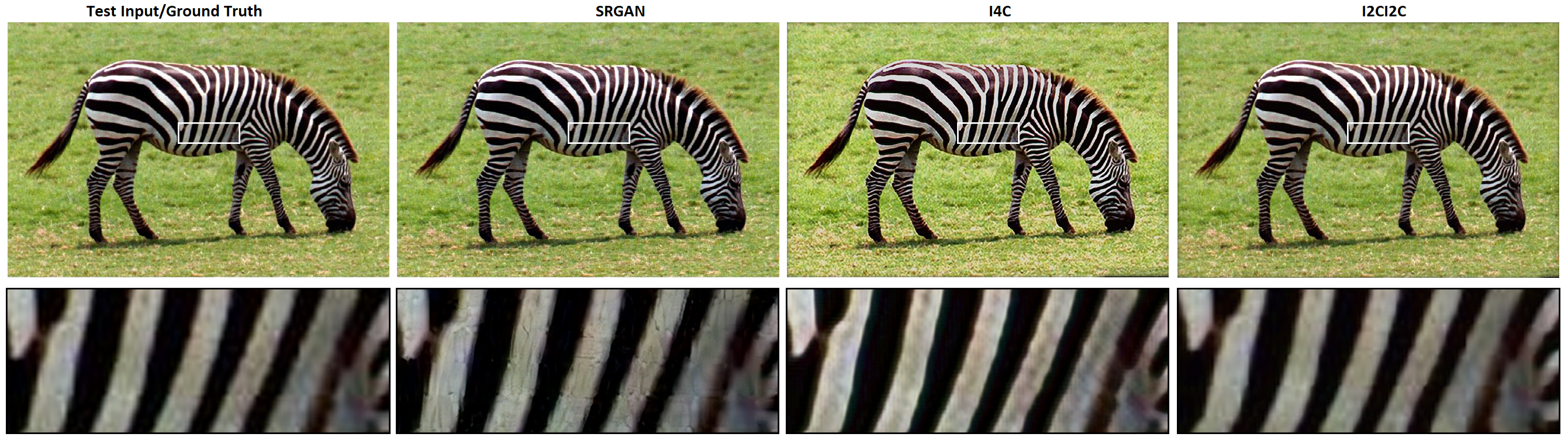}
		\caption{Comparison of the results of our 4X upscaling methods (I4C and I2CI2C) with that of SRGAN, when the original zebra image from test-set set14 is directly input to the models. The output of SRGAN is sharper than the original, while the output of I2CI2C has a texture closest to the original.}
		\label{results_4_fig_1}
	\end{figure*}

\begin{figure*}
		%\centering
		\includegraphics[width=0.98\textwidth,height=0.20\textheight]{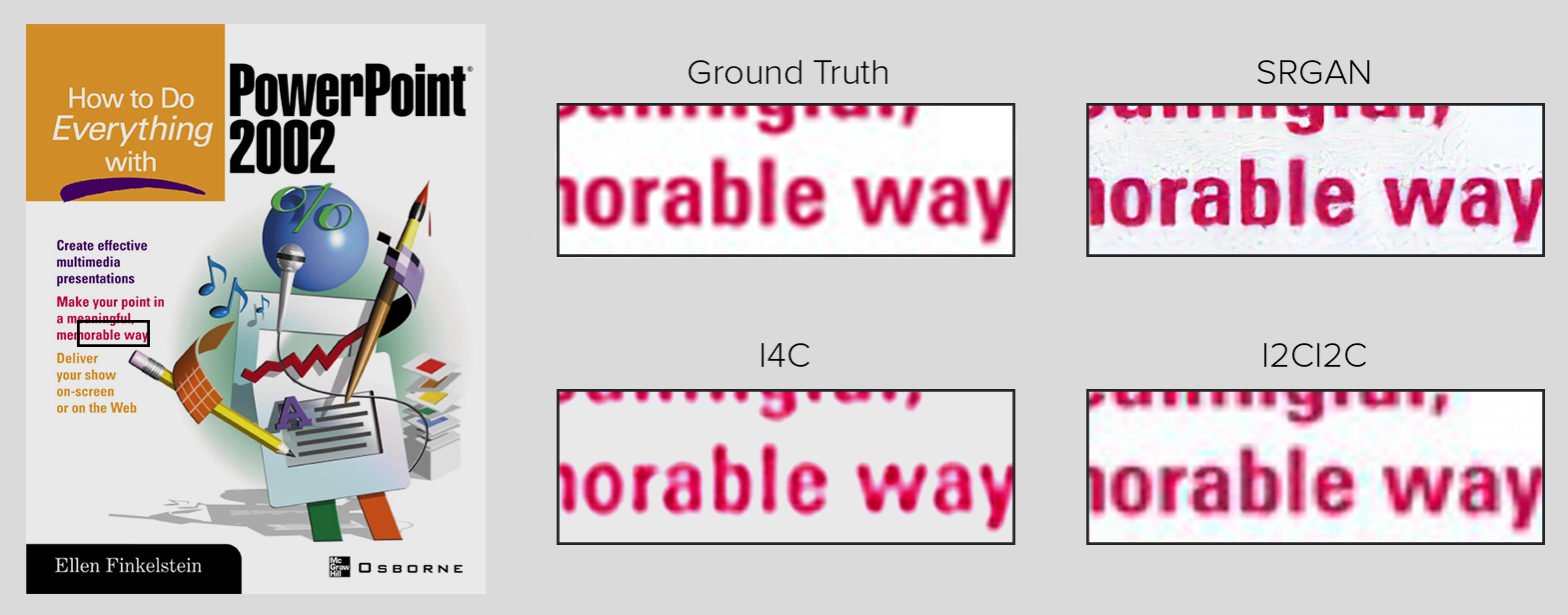}
		\caption{Comparison of the results of our 4X upscaling methods (I4C and I2CI2C) with that of SRGAN, when the original ppt3 image from test-set set14 is directly input to the models. The output of I4C is somewhat comparable to the original image, but marginally less sharp. SRGAN delivers an output sharper than the original, but also introduces some speckle/pepper noise.}
		\label{results_4_fig_2}
	\end{figure*}

	Throughout this work, we have focused on upscaling the images by a factor of 2. However, we used similar ideas to also develop two main architectures (I4C and I2CI2C) for upscaling natural images by a factor of 4. These architectures, shown in Figs. \ref{arch_4_1} and \ref{arch_4_2}, have been developed keeping in mind that they should generalize well for almost all types of datasets. We find that our methods generate output images of good perceptual quality, when we directly pass the ground truth image as the test image to our model. Figures~\ref{results_4_fig_1} and~\ref{results_4_fig_2} compare the results of I4C and I2CI2C methods with that of SRGAN~\cite{srgan}, which is a recently proposed, deep learning based technique, which claims that it reconstructs the image precisely in the manifold of the HR image. The perceptual quality of our results is as good as (or better than) that of SRGAN, which uses millions of parameters to reconstruct the HR image. Thus, it shows that good results are possible with architectures less complex than architectures like SRGAN.
	
	\section{Conclusion}
	Our approach is unique in creating a diverse input space using the simple techniques of multiple interpolations and then non-linearly combining them by a CNN to obtain high resolution images. Unlike most methods in the literature, we use all the three color channels together to train and test our models. We have proposed four main algorithms, two each for upscaling factors of two and four. The first one uses a deep architecture to combine the outputs of nearest neighbor, bilinear and bicubic interpolations. The second architecture uses skip connection by nearest neighbor interpolation. Our focus has been on the generalization of the performance across diverse datasets and both of these techniques perform reasonably well. The proposed architectures use less number of parameters than most of the recent state-of-the-art techniques in the literature and hence are computationally efficient.
	Our CI2 and CB2SNN methods outperform most of the techniques in terms of the PSNR values, when the complexity and size of the dataset increases. The nearest neighbor interpolation, which has been largely ignored by the deep learning community, plays an important role in our architectures. Interpolation by various techniques can be used to widen the input image space. This results in a more efficient architecture, that generalizes well for most datasets. The perceptual quality of our 4X upscaling result is comparable to that of the recently reported technique SRGAN. With a few exceptions, our results are comparable to the most recently reported state of the art methods based on deep learning.
	\section{Future work}
	We have carried out a large number of experiments on different structures and parameter settings to strike a balance between the quality of the output image and the time taken to obtain the same. Incorporating additional techniques to extract other distinctive features before passing them to a CNN may be a better algorithm to improve quality. A mathematical study of the proposed algorithms may help improve their performance further. Other interpolation techniques can be explored and an optimal weighting of other interpolations may also lead to better results.
	
	\section{Acknowledgment}
	The authors thank Mr. Aswin Vasan for helping in creating the figures used in this paper.

\end{document}